\pdfoutput=1
\documentclass[11pt]{article}

\usepackage{emnlp2021}

\usepackage{times}
\usepackage{graphicx}
\usepackage{latexsym}
\usepackage{amsfonts}
\usepackage{amsmath}
\usepackage{multirow}
\usepackage{latexsym}
\usepackage{paralist}
\usepackage{booktabs}
\usepackage[T1]{fontenc}
\usepackage[utf8]{inputenc}

\usepackage{microtype}
\title{Heterogeneous Graph Neural Networks for Keyphrase Generation}

\author{
Jiacheng Ye$^{1}$\thanks{$^*$ Equal contribution.}, 
Ruijian Cai$^{1^*}$, 
Tao Gui$^{2}$\thanks{$^\dag$ Corresponding authors.} and 
Qi Zhang$^{1^\dag}$ \\
$^1$School of Computer Science, Shanghai Key Laboratory of Intelligent Information Processing, \\
Fudan University, Shanghai, China \\
$^2$Institute of Modern Languages and Linguistics, Fudan University \\
{\tt \{yejc19, 19210240253, tgui, qz\}@fudan.edu.cn}
}

\begin{document}

\maketitle

\begin{abstract}
The encoder-decoder framework achieves state-of-the-art results in keyphrase generation (KG) tasks by predicting both present keyphrases that appear in the source document and absent keyphrases that do not. However, relying solely on the source document can result in generating uncontrollable and inaccurate absent keyphrases.
To address these problems, we propose a novel graph-based method that can capture explicit knowledge from related references. Our model first retrieves some document-keyphrases pairs similar to the source document from a pre-defined index as references. Then a heterogeneous graph is constructed to capture relationships of different granularities between the source document and its references. To guide the decoding process, a hierarchical attention and copy mechanism is introduced, which directly copies appropriate words from both the source document and its references based on their relevance and significance. 
The experimental results on multiple KG benchmarks show that the proposed model achieves significant improvements against other baseline models, especially with regard to the absent keyphrase prediction.
\end{abstract}

\section{Introduction}
Keyphrase generation (KG), a fundamental task in the field of natural language processing (NLP), refers to the generation of a set of keyphrases that expresses the crucial semantic meaning of a document. These keyphrases can be further categorized into present keyphrases that appear in the document and absent keyphrases that do not. Current KG approaches generally adopt an encoder-decoder framework \cite{sutskever2014sequence} with attention mechanism \cite{bahdanau2014neural,luong-etal-2015-effective} and copy mechanism \cite{gu2016incorporating,see-etal-2017-get} to simultaneously predict present and absent keyphrases \cite{meng2017,chen2018a,chan2019,chen2019a,chen2019,yuan2018}.

\begin{figure}[t]
\centering
\includegraphics[width=3in]{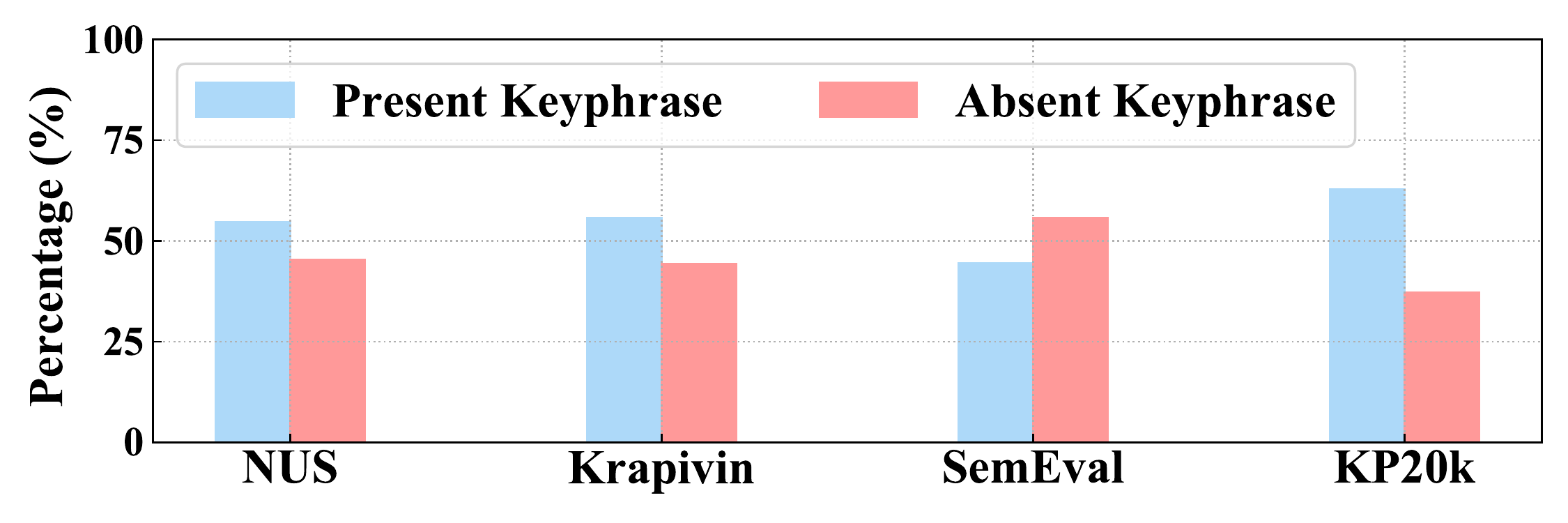}
\caption{Proportion of present and absent keyphrases among four datasets. Although the previous methods for keyphrase generation have shown promising results on present keyphrase predictions, they are not yet satisfactory on the absent keyphrase predictions, which also occupy a large proportion.}
\label{fig:gap}
\end{figure}

Although the proposed methods for keyphrase generation have shown promising results on present keyphrase predictions, 
% they are not yet satisfactory on the absent ones. 
they often generate uncontrollable and inaccurate predictions on the absent ones.
% The main reason is that there are numerous \textit{implicit} relationships, between absent keyphrases and concepts in the document (e.g., technology hypernyms or task hypernyms). 
The main reason is that there are numerous candidates of absent keyphrases that have implicit relationships (e.g., technology hypernyms or task hypernyms) with the concepts in the document. For instance, for a document discussing ``LSTM'', all the technology hypernyms like ``Neural Network'', ``RNN'' and ``Recurrent Neural Network'' can be its absent keyphrases candidates. 
% For a document about ``sentiment analysis'', ``text classification'' and ``NLP'' as task hypernyms can also be its absent candidates. 
When dealing with scarce training data or limited model size, it is non-trivial for the model to summarize and memorize all the candidates accurately. 
Thus, 
one can expect that the generated absent keyphrases are often sub-optimal when the candidate set in model’s mind is relatively small or inaccurate.
% the generated absent keyphrases are often uncontrollable and inaccurate when relying solely on the source document.
% the candidate set in model's mind is relatively small or inaccurate.
This problem is crucial because absent keyphrases account for a large proportion of all the ground-truth keyphrases. As shown in Figure \ref{fig:gap}, in some datasets, up to 50\% of the keyphrases are absent. 

\begin{figure*}[t]
\centering
\includegraphics[width=6in]{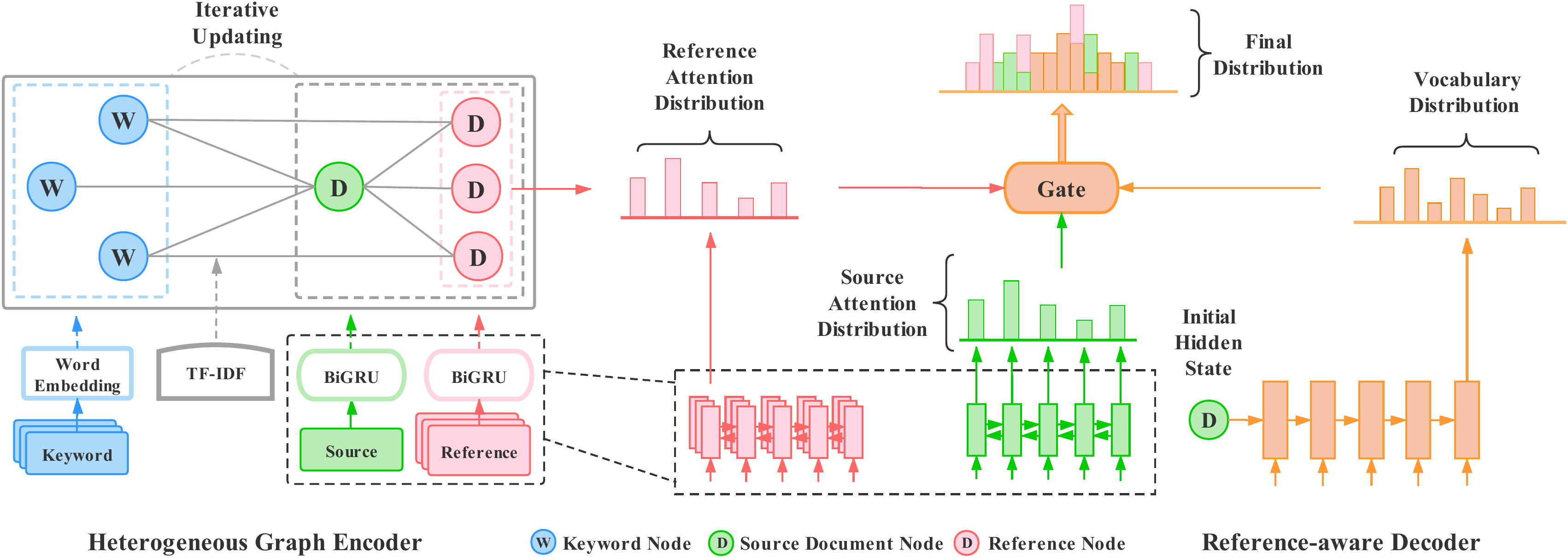}
\caption{Graphical illustration of our proposed \textsc{Gater}. We first retrieve references using the source document, where each reference is the concatenation of document and keyphrases pair from the training set. Then we construct a heterogeneous graph and perform iterative updating. Finally, the source document node is extracted to decode the keyphrase sequence with a hierarchical attention and copy mechanism.}
\label{fig:model}
\end{figure*}

To address this problem, we propose a novel graph-based method to capture explicit knowledge from related references. Each reference is a retrieved document-keyphrases pair from a predefined index (e.g., the training set) that similar to the source document. 
This is motivated by the fact that the related references often contain candidate or even ground-truth absent keyphrases of the source document. Empirically, we find three retrieved references cover up to 27\% of the ground-truth absent keyphrases on average (see Section \ref{section:references} for details). 

Our heterogeneous graph is designed to incorporate knowledge from the related references. It contains source document, reference and keyword nodes, and has the following advantages: (a) different reference nodes can interact with the source document regarding the explicit shared keyword information, which can enrich the semantic representation of the source document; (b) a powerful structural prior is introduced as the keywords are highly overlapped with the ground-truth keyphrases. Statistically, we collect the top five keywords from each document on the validation set, and we find that these keywords contain 68\% of the tokens in the ground-truth keyphrases. 
On the decoder side, as a portion of absent keyphrases directly appear in the references, we propose a hierarchical attention and copy mechanism for copying appropriate words from both source document and its references based on their relevance and significance.

The main contributions of this paper can be summarized as follows: (1) we design a heterogeneous graph network for keyphrase generation, which can enrich the source document node through keyword nodes and retrieved reference nodes; 
(2) we propose a hierarchical attention and copy mechanism to facilitate the decoding process, which can copy appropriate words from both the source document and retrieved references; and (3) our proposed method outperforms other state-of-the-art methods on multiple benchmarks, and especially excels in absent keyphrase prediction. Our codes are publicly available at \textit{Github}\footnote{\url{https://github.com/jiacheng-ye/kg_gater}}.

\section{Methodology}
In this work, we propose a heterogeneous Graph ATtention network basEd on References (\textsc{Gater}) for keyphrase generation, as shown in Figure \ref{fig:model}. Given a source document, we first retrieve related document from a predefined index\footnote{We use the training set as our reference index in our experiment, which can also be easily extended to open corpus.} and concatenate each retrieved document with its keyphrases to serve as a reference. Then we construct a heterogeneous graph that contains document nodes\footnote{Note that source document and reference are the two specific contents of the document node.} and keyword nodes based on the source document and its references. The graph is updated iteratively to enhance the representations of the source document node. Finally, the source document node is extracted to decode the keyphrase sequence. To facilitate the decoding process, we also introduce a hierarchical attention and copy mechanism, with which the model directly attends to and copies from both the source document and its references. The hierarchical arrangement ensures that more semantically relevant words and those in more relevant references will be given larger weights for the current decision. 

\subsection{Reference Retriever}
Given a source document $\mathbf{x}$, we first use a reference retriever to output several related references from the training set. To make full use of both the retrieved document and retrieved keyphrases, we denote a reference as the concatenation of the two. We find that the use of a term frequency–inverse document frequency (TF-IDF)-based retriever provides a simple but efficient means to accomplish the retrieval task. Specifically, we first represent the source document and all the reference candidates as TF-IDF weighted uni/bi-gram vectors. Then, the most similar $K$ references $\mathcal{X}^r=\{\mathbf{x}^{r_i}\}_{i=1,\dots,K}$ are retrieved by comparing the cosine similarities of the vectors of the source document and all the references.

\subsection{Heterogeneous Graph Encoder}
\subsubsection{Graph Construction}
Given the source document $\mathbf{x}$ and its references $\mathcal{X}^r$, we select the top-$k$ unique words as keywords based on their TF-IDF weights from the source document and each reference. The additional keyword nodes can enrich the semantic representation of the source document through message passing, and introduce prior knowledge for generating keyphrase as the highly overlap between keywords and keyphrases. 
We then build a heterogeneous graph based on the source document, references and keywords.

Formally, our undirected heterogeneous graph can be defined as $G=\{V, E\}$, $V=V_w \cup V_d$ and $E=E_{d2d} \cup E_{w2d}$. Specifically, $V_w=\{w_i\}$ ($i \in \{1,\dots,m\}$) denotes $m$ unique keyword nodes of the source document and $K$ references, $V_d=\mathbf{x} \cup \mathcal{X}^r$ corresponds to the source document node and $K$ reference nodes, $E_{d2d}=\{e_k\}$ ($k \in \{1,\dots,K\}$) and $e_{k}$ represents the edge weight between the $k$-th reference and source document, and $E_{w2d}=\{e_{i,j}\}$ ($i \in \{1,\dots,m\},j\in\{1,\dots,K+1\}$) and $e_{i,j}$ indicates the edge weight between the $i$-th keyword and the $j$-th document.

\subsubsection{Graph Initializers}
\paragraph{Node Initializers}
There are two types of nodes in our heterogeneous graph (i.e., document nodes $V_d$ and keyword nodes $V_w$). For each document node, the same as previous works \cite{meng2017,chen2019}, an embedding lookup table $\mathbf{e}^w$ is first applied to each word, and then a bidirectional Gated Recurrent Unit (GRU) \cite{cho2014learning} is used to obtain the context-aware representation of each word. The representation for document $\mathbf{x}$ and each word is defined as the concatenation of the forward and backward hidden states (i.e., $\mathbf{d}=[\overrightarrow{\mathbf{m}}_1;\overleftarrow{\mathbf{m}}_{L_\mathbf{x}}]$ and $\mathbf{m}_i=[\overrightarrow{\mathbf{m}}_i;\overleftarrow{\mathbf{m}}_i]$, respectively). 
For each keyword node, since the same keyword may appear in multiple documents, we simply use the word embedding as its initial node representation $\mathbf{w}_i=\mathbf{e}^w(w_i)$.

\paragraph{Edge Initializers}
There are two types of edges in our heterogeneous graph (i.e., document-to-document edge $E_{d2d}$ and document-to-keyword $E_{d2w}$). To include information about the significance of the relationships between keyword and document nodes, we infuse TF-IDF values in the edge weights. Similarly, we also infuse TF-IDF values in the edge weights of $E_{d2d}$ as a prior statistical $n$-gram similarity between documents. The two types of floating TF-IDF weights are then transformed into integers and mapped to dense vectors using embedding matrices $\mathbf{e}^{d2d}$ and $\mathbf{e}^{w2d}$.

\subsubsection{Graph Aggregating and Updating}
\paragraph{Aggregator}

Graph attention networks (GAT) \cite{velivckovic2018graph} are used to aggregate information for each node. We denote the hidden states of input nodes as $\mathbf{h}_i \in \mathbb{R}^{d_h}$, where $i \in \{1,\dots,N\}$. With the additional edge feature, the aggregator is defined as follows:

\begin{equation}
\begin{aligned}
z_{i j}&=\text{LeakyReLU}\left(\mathbf{w}_{a}^T\left[\mathbf{W}_{q} \mathbf{h}_{i} ; \mathbf{W}_{k} \mathbf{h}_{j}; \mathbf{e}_{ij}\right]\right) \\
\alpha_{i j}&=\operatorname{softmax}_{j}\left(z_{i j}\right)=\frac{\exp \left(z_{i j}\right)}{\sum_{k \in \mathcal{N}_{i}} \exp \left(z_{i k}\right)} \\
\mathbf{u}_{i}&=\sigma(\sum_{j \in \mathcal{N}_{i}} \alpha_{i j} \mathbf{W}_{v} \mathbf{h}_{j}),
\end{aligned}
\end{equation}
where $\mathbf{e}_{ij}$ is the embedding of edge feature, $\alpha_{i j}$ is the attention weight between $\mathbf{h_i}$ and $\mathbf{h_j}$, and $\mathbf{u}_{i}$ is the aggregated feature. For simplicity, we will use $\operatorname{GAT}\left(\mathbf{H}, \mathbf{H}, \mathbf{H}, \mathbf{E} \right)$ to denote the GAT aggregating layer, where $\mathbf{H}$ is used for query, key, and value, and $\mathbf{E}$ is used as edge features.

\paragraph{Updater}
To update the node state, similar to the approach used in the Transformer \cite{vaswani2017attention}, we introduce a residual connection and position-wise feed-forward (FFN) layer consisting of two linear transformations. Given an undirected heterogeneous graph $G$ with node features $\mathbf{H}_w \cup \mathbf{H}_d$ and edge features $\mathbf{E}_{w2d} \cup \mathbf{E}_{d2d}$, we update each types of nodes separately as follows:

\begin{equation}
\begin{aligned}
\mathbf{H}^{1}_w &=\mathrm{FFN}\left(\operatorname{GAT}\left(\mathbf{H}^{0}_w, \mathbf{H}^{0}_d, \mathbf{H}^{0}_d, \mathbf{E}_{w2d} \right)+\mathbf{H}^{0}_w \right)\\
\mathbf{H}^{1}_d &=\mathrm{FFN}\left(\operatorname{GAT}\left(\mathbf{H}^{0}_d, \mathbf{H}^{1}_w, \mathbf{H}^{1}_w, \mathbf{E}_{w2d} \right)+\mathbf{H}^{0}_d \right)\\
\mathbf{H}^{1}_d &=\mathrm{FFN}\left(\operatorname{GAT}\left(\mathbf{H}^{1}_d, \mathbf{H}^{1}_d, \mathbf{H}^{1}_d, \mathbf{E}_{d2d} \right)+\mathbf{H}^{1}_d \right),
\end{aligned}
\end{equation}
with word nodes updated first by aggregating document-level information from document nodes, then document nodes updated by the updated word nodes, and finally document nodes updated again by the updated document nodes. The above process is executed iteratively for $I$ steps to realize better document representation. 

When the heterogeneous graph encoder finished, we seperate $\mathbf{H}^{I}_d$ into $\mathbf{d}^s$ and $\mathbf{D}^{r}=\{{\mathbf{d}}^{r_i}\}_{i=1\dots, K}$ as the representation of source document and each reference. We denote $\mathbf{M}^s=\{\mathbf{m}^s_{i}\}_{i=1,\dots,L_{\mathbf{x}}}$ as the encoder hidden state of each word in the source document, $\mathbf{M}^{r}=\{\mathbf{M}^{r_i}\}_{i=1\dots, K}$ and $\mathbf{M}^{r_i}=\{\mathbf{m}^{r_i}_{j}\}_{j=1\dots, L_{r_i}}$ denotes the encoder hidden state of each word of the $i$-th reference. 
All the features described above (i.e., $\mathbf{d}^s$, $\mathbf{D}^{r}$, $\mathbf{M}^s$ and $\mathbf{M}^{r}$) will be used in the reference-aware decoder.

\subsection{Reference-aware Decoder}
After encoding the document into a reference-aware representation $\mathbf{d}^s$, we propose a hierarchical attention and copy mechanism to further incorporate the reference information by attending to and copying words from both the source document and the references.

% \paragraph{Hierarchical Attention Mechanism}
We use $\mathbf{d}^s$ as the initial hidden state of a GRU decoder, and the decoding process in time step $t$ is described as follows:
\begin{equation}
\begin{aligned}
% \begin{array}{l}
\mathbf{h}_{t} &=\operatorname{GRU}(\mathbf{e}^w(y_{t-1}), \mathbf{h}_{t-1}) \\
\mathbf{c}_{t} &=\operatorname{hier\_attn}(\mathbf{h}_{t},\mathbf{M}^s,\mathbf{M}^{r},\mathbf{D}^{r}) \\
\tilde{\mathbf{h}}_{t} &=\tanh (\mathbf{W}_c[\mathbf{c}_{t} ; \mathbf{h}_{t}]), \\
% \end{array}
\end{aligned}
\end{equation}
where $\mathbf{c}_{t}$ is the context vector and the hierarchical attention mechanism $\operatorname{hier\_attn}$ is defined as follows: 
\begin{equation}
\begin{aligned}
% \begin{array}{l}
\mathbf{c}_{t}^{s} &= \sum_{i=1}^{L_{\mathbf{x}}} {a^s_{t,i}}\mathbf{m}^s_i ;
\mathbf{c}^r_{t} = \sum_{i=1}^K {\sum_{j=1}^{L_{\mathbf{x}^{r_i}}}} {a^{r}_{t,i}} a_{t,j}^{r_i} \mathbf{m}^{r_i}_j \\
\mathbf{c}_{t} &= g_{ref} \cdot \mathbf{c}^s_{t} + (1-g_{ref}) \cdot \mathbf{c}^r_{t},
% \end{array}
\end{aligned}
\end{equation}
where $\mathbf{a}^s_{t}$ is a word-level attention distribution over words from the source document using $\mathbf{M}^s$, $\mathbf{a}_t^{r}$ is an attention distribution over references using $\mathbf{D}^{r}$, which gives greater weights to more relevant references, $\mathbf{a}_{t}^{r_i}$ is a word-level attention distribution over words from $i$-th reference using $\mathbf{M}^{r_i}$, which can be considered as the importance of each word in $i$-th reference, and $g_{ref} = \operatorname{sigmoid}(\mathbf{w}_{ref}[\mathbf{c}^s_{t};\mathbf{c}^r_{t}])$ is a soft gate for determining the importance of the context vectors from source document and references. All the attention distributions described above are computed as in \citet{bahdanau2014neural}.
% , which is defined as follows:
% \begin{equation}
% \begin{aligned}
% % \begin{array}{l}
% e_{t,i}&=v^{T} \tanh \left(W_{a} [\mathbf{h}_{t};\mathbf{m}_{i}] \right) \\
% \mathbf{a}_{t}&=\operatorname{softmax}\left(\mathbf{e}_{t}\right).
% \label{eq:attn}
% % \end{array}
% \end{aligned}
% \end{equation}

% \paragraph{Hierarchical Copy Mechanism}
To alleviate the out-of-vocabulary (OOV) problem, a copy mechanism \cite{see-etal-2017-get} is generally adopted. To further guide the decoding process by copying appropriate words from references based on their relevance and significance, we propose a hierarchical copy mechanism. Specifically, a dynamic vocabulary $\mathcal{V}^{\prime}$ is constructed by merging the predefined vocabulary $\mathcal{V}$, the words in source document $\mathcal{V}_\mathbf{x}$ and all the words in the references $\mathcal{V}_{\mathcal{X}^r}$. Thus, the probability of predicting a word $y_t$is computed as follows:
\begin{equation}
P_{\mathcal{V}^{\prime}}\left(y_{t}\right)=p_1 P_{\mathcal{V}}\left(y_{t}\right)+p_2 P_{\mathcal{V}_\mathbf{x}}\left(y_{t}\right) +p_3 P_{\mathcal{V}_{\mathcal{X}^r}}\left(y_{t}\right),
\end{equation}
where $P_{\mathcal{V}}(y_{t})=\operatorname{softmax}(\operatorname{MLP}([\mathbf{h}_{t} ; \tilde{\mathbf{h}}_{t}]))$ is the generative probability over predefined vocabulary $\mathcal{V}$,
 $P_{\mathcal{V}_\mathbf{x}}\left(y_{t}\right)=\sum_{i: x_{i}=y_{t}} a^s_{t, i}$ is the copy probability from the source document,
 $P_{\mathcal{V}_{\mathcal{X}^r}}(y_t)=\sum_i\sum_{j:x_j^{r_i}=y_t} a^{r_i}_{t,j}$ is the copy probability from all the references, and 
$\mathbf{p}=\operatorname{softmax}(\mathbf{W}_p[\tilde{\mathbf{h}}_{t}; \mathbf{h}_{t} ; \mathbf{e}^w(y_{t-1})]) \in \mathbb{R}^3$ serves as a soft switcher that determines the preference for selecting the word from the predefined vocabulary, source document or references.

\subsection{Training}
The proposed \textsc{Gater} model is independent of any specific training method, so we can use either the \textsc{One2One} training paradigm \cite{meng2017}, where the target keyphrase set $\mathcal{Y}=\{\mathbf{y}_i\}_{i=1,\dots,|\mathcal{Y}|}$ are split into multiple training targets for a source document $\mathbf{x}$: 
% \begin{small}
\begin{equation}\small
% \begin{array}{l}
\mathcal{L}_{\text{\textsc{One2One}}}(\theta)=-\sum_{i=1}^{|\mathcal{Y}|} \sum_{t=1}^{L_{\mathbf{y}_i}} \log P_{\mathcal{V}^{\prime}}\left(y_{i,t} \mid \mathbf{y}_{i,1:t-1}, \mathbf{x}; \theta\right), 
% \end{array}
\end{equation}
% \end{small}
or the \textsc{One2Seq} training paradigm \cite{ye2018,yuan2018}, where all the keyphrases are concatenated into one training target:
\begin{equation}
% \begin{array}{l}
\mathcal{L}_{\textsc{One2Seq}}(\theta)=-\sum_{t=1}^{L_{\mathbf{y^\star}}} \log P_{\mathcal{V}^{\prime}}\left(y_{t}^\star \mid \mathbf{y^\star}_{1:t-1}, \mathbf{x}; \theta\right),
% \end{array}
\end{equation}
where $\mathbf{y^\star}$ is the concatenation of the keyphrases in $\mathcal{Y}$ by a delimiter.

\begin{table*}[t]
\centering
\scalebox{0.66}{
% \begin{tabular}{l|cccc|cccc|cccc}
\begin{tabular}{l|llll|llll|llll}
\toprule
\multicolumn{1}{l|}{\multirow{3}{*}{\textbf{Model}}} & \multicolumn{4}{c|}{\textbf{NUS}} & \multicolumn{4}{c|}{\textbf{SemEval}} & \multicolumn{4}{c}{\textbf{KP20k}} \\

& \multicolumn{2}{c}{\textbf{Present}} & \multicolumn{2}{c|}{\textbf{Absent}} & \multicolumn{2}{c}{\textbf{Present}} & \multicolumn{2}{c|}{\textbf{Absent}} & \multicolumn{2}{c}{\textbf{Present}} & \multicolumn{2}{c}{\textbf{Absent}} \\

\multicolumn{1}{c|}{} & {$F1@5$} & $F1@10$ & $R@10$ & $R@50$ & $F1@5$ & $F1@10$ & $R@10$ & $R@50$ & $F1@5$ & $F1@10$ & $R@10$ & $R@50$ \\
\hline
CopyRNN \cite{meng2017} & 0.311 & 0.266 & 0.058 & 0.116 & 0.293 & 0.304 & 0.043 & 0.067 & 0.333 & 0.262 & 0.125 & 0.211 \\
CorrRNN \cite{chen2018a} & 0.318 & 0.278 & 0.059 & - & 0.320 & 0.320 & 0.041 & - & - & - & - & - \\
TG-Net \cite{chen2019a} & 0.349 & 0.295 & 0.075 & 0.137 & 0.318 & 0.322 & 0.045 & 0.076 & 0.372 & 0.315 & 0.156 & 0.268 \\
KG-KE-KR-M \cite{chen2019} & 0.344 & 0.287 & 0.123 & \textbf{0.193} & 0.329 & 0.327 & 0.049 & 0.090 & 0.400 & \textbf{0.327} & 0.177 & 0.278 \\
\hline
CopyRNN-\textsc{Gater} (Ours) & \textbf{0.374$_4$} & \textbf{0.304$_4$} & \textbf{0.126$_3$} & \textbf{0.193$_2$} & \textbf{0.366$_3$} & \textbf{0.340$_4$} & \textbf{0.056$_1$} & \textbf{0.092$_2$} & \textbf{0.402$_1$} & 0.324$_1$ & \textbf{0.186$_0$} & \textbf{0.285$_1$} \\
% \% gain & 9\% & 6\% & 3\% & 0\% & 11\% & 4\% & 14\% & 2\% & 1\% & -1\% & 5\% & 3\% \\
\bottomrule
  \end{tabular}}
\caption{Keyphrase prediction results of all the models trained under \textsc{One2One} paradigm. The best results are bold. The subscript are corresponding standard deviation (e.g., 0.285$_1$ means 0.285$\pm$0.001). 
% ``\% gain'' refers to the improvement gain over previous best model (i.e., KG-KE-KR-M).
}
\label{tab:one2one}
\end{table*}

\begin{table*}[t]
\centering
\scalebox{0.66}{
% \begin{tabular}{l|cccc|cccc|cccc}
\begin{tabular}{l|llll|llll|llll}
\toprule
\multicolumn{1}{l|}{\multirow{3}{*}{\textbf{Model}}} & \multicolumn{4}{c|}{\textbf{NUS}} & \multicolumn{4}{c|}{\textbf{SemEval}} & \multicolumn{4}{c}{\textbf{KP20k}} \\

& \multicolumn{2}{c}{\textbf{Present}} & \multicolumn{2}{c|}{\textbf{Absent}} & \multicolumn{2}{c}{\textbf{Present}} & \multicolumn{2}{c|}{\textbf{Absent}} & \multicolumn{2}{c}{\textbf{Present}} & \multicolumn{2}{c}{\textbf{Absent}} \\

\multicolumn{1}{c|}{} & $F1@5$ & $F1@M$ & $F1@5$ & $F1@M$ & $F1@5$ & $F1@M$ & $F1@5$ & $F1@M$ & $F1@5$ & $F1@M$ & $F1@5$ & $F1@M$ \\
 \hline
catSeq \cite{yuan2018} & 0.323 & 0.397 & 0.016 & 0.028 & 0.242 & 0.283 & 0.020 & 0.028 & 0.291 & 0.367 & 0.015 & 0.032 \\
catSeqD \cite{yuan2018}& 0.321 & 0.394 & 0.014 & 0.024 & 0.233 & 0.274 & 0.016 & 0.024 & 0.285 & 0.363 & 0.015 & 0.031 \\
catSeqCorr \cite{chan2019} & 0.319 & 0.390 & 0.014 & 0.024 & 0.246 & 0.290 & 0.018 & 0.026 & 0.289 & 0.365 & 0.015 & 0.032 \\
catSeqTG \cite{chan2019} & 0.325 & 0.393 & 0.011 & 0.018 & 0.246 & 0.290 & 0.019 & 0.027 & 0.292 & 0.366 & 0.015 & 0.032 \\
SenSeNet \cite{luo2020sensenet} & \textbf{0.348} & 0.403 & 0.018 & 0.032 & 0.255 & 0.299 & 0.024 & 0.032 & \textbf{0.296} & 0.370 & 0.017 & 0.036 \\
\hline
catSeq-\textsc{Gater} (Ours) & 0.337$_4$ & \textbf{0.418$_4$} & \textbf{0.033$_3$} & \textbf{0.054$_4$} & \textbf{0.257$_3$} & \textbf{0.309$_4$} & \textbf{0.026$_4$} & \textbf{0.035$_5$} & 0.295$_2$ & \textbf{0.384$_1$} & \textbf{0.030$_1$} & \textbf{0.060$_2$} \\
% \% gain & -3\% & 4\% & 136\% & 122\% & 1\% & 3\% & 11\% & 11\% & 0\% & 4\% & 87\% & 75\% \\
\bottomrule
\end{tabular}}
\caption{Keyphrase prediction results of all the models trained under \textsc{One2Seq} paradigm. The best results are bold. The subscript are corresponding standard deviation (e.g., 0.060$_2$ means 0.060$\pm$0.002). 
% ``\% gain'' refers to the improvement gain over previous best model (i.e., SenSeNet). 
}
\label{tab:one2seq}
\end{table*}

\section{Experimental Setup}
\subsection{Datasets}
We conduct our experiments on four scientific article datasets, including \textbf{NUS} \cite{nguyen2007keyphrase}, \textbf{Krapivin} \cite{krapivin2009large}, \textbf{SemEval} \cite{kim2010semeval} and \textbf{KP20k} \cite{meng2017}. 
Each sample from these datasets consists of a title, an abstract, and some keyphrases given by the authors of the papers. Following previous works \cite{meng2017,chen2019a,chen2019,yuan2018}, we concatenate the title and abstract as a source document. 
We use the largest dataset (i.e., \textbf{KP20k}) for model training, and the testing sets of all the four datasets for evaluation. After preprocessing (i.e., lowercasing, replacing all the digits with the symbol $\langle\mathit{digit}\rangle$ and removing the duplicated data), the final \textbf{KP20k} dataset contains 509,818 samples for training, 20,000 for validation and 20,000 for testing. The number of test samples in \textbf{NUS}, \textbf{Krapivin} and \textbf{SemEval} is 211, 400 and 100, respectively.

\subsection{Baselines}
For a comprehensive evaluation, we verify our method under both training paradigms (i.e., \textsc{One2One} and \textsc{One2Seq}) and compare with the following methods\footnote{We didn't compare with \citet{chen2020a} since they use a different preprocessing method with others, see \href{https://github.com/Chen-Wang-CUHK/ExHiRD-DKG/issues/7\#issuecomment-681317389}{the discussion on github} for details.}:

\begin{compactitem}
\item \textbf{catSeq} \cite{yuan2018}. The RNN-based seq2seq model with copy mechanism under \textsc{One2Seq} training paradigm. 
\textbf{CopyRNN} \cite{meng2017} is the one with the same model but under \textsc{One2One} training paradigm.

\item \textbf{catSeqD} \cite{yuan2018}. An extension of catSeq with orthogonal regularization \cite{bousmalis2016domain} and target encoding to improve diversity under \textsc{One2Seq} training paradigm.

\item \textbf{catSeqCorr} \cite{chan2019}. The extension of catSeq with coverage and review mechanisms under \textsc{One2Seq} training paradigm. 
\textbf{CorrRNN} \cite{chen2018a} is the one under \textsc{One2One} training paradigm.

\item \textbf{catSeqTG} \cite{chan2019}. The extension of catSeq with additional title encoding. 
\textbf{TG-Net} \cite{chen2019a} is the one under \textsc{One2One} training paradigm.

\item \textbf{KG-KE-KR-M} \cite{chen2019}. A joint extraction and generation model with the retrieved keyphrases and a merging process under \textsc{One2One} training paradigm.

\item \textbf{SenSeNet} \cite{luo2020sensenet}. The extension of catSeq with document structure under \textsc{One2Seq} paradigm.

\end{compactitem}

\subsection{Implementation Details}
Following previous works \cite{chan2019,yuan2018}, when training under the \textsc{One2Seq} paradigm, the target keyphrase sequence is the concatenation of present and absent keyphrases, with the present keyphrases are sorted according to the orders of their first occurrences in the document and the absent keyphrase kept in their original order.

We keep all the parameters the same as those reported in \citet{chan2019}, hence, we only report the parameters in the additional graph module. 
We retrieve 3 references and extract the top 20 keywords from source document and each reference to construct the graph. We set the number of attention heads to 5 and the number of iterations to 2, based on the valid set. 
During training, we use a dropout rate of 0.3 for the graph layer, the batch size of 12 and 64 for \textsc{One2Seq} and \textsc{One2One} training paradigm, respectively.
During testing, we use greedy search for \textsc{One2Seq}, and beam search with a maximum depth of 6 and a beam size of 200 for \textsc{One2One}. 
We repeat the experiments of our model three times using different random seeds and report the averaged results.

\subsection{Evaluation Metrics}
For the model trained under \textsc{One2One} paradigm, as in previous works \cite{meng2017,chen2018a,chen2019a}, we use macro-averaged $F_1@5$ and $F_1@10$ for present keyphrase predictions, and $R@10$ and $R@50$ for absent keyphrase predictions. 
For the model trained under \textsc{One2Seq} paradigm, we follow \citet{chan2019} and use $F_1@5$ and $F_1@M$ for both present and absent keyphrase predictions, where $F_1@M$ compares all the keyphrases predicted by the model with the ground-truth keyphrases, which means it considers the number of predictions. 
We apply the Porter Stemmer before determining whether two keyphrases are identical and remove all the duplicated keyphrases after stemming.

\section{Results and Analysis}
\subsection{Present and Absent Keyphrase Predictions}

Table \ref{tab:one2one} and Table \ref{tab:one2seq} show the performance evaluations of the present and absent keyphrase predicted by the model trained under \textsc{One2One} paradigm and \textsc{One2Seq} paradigm, respectively.\footnote{Due to the space limitations, the results on the Krapivin dataset can be found in Appendix \ref{sec:appendix}.}
For the results on absent keyphrases, as noted by previous works \cite{chan2019,yuan2018} that predicting absent keyphrases for a document is an extremely challenging task, the proposed \textsc{Gater} model still outperforms the state-of-the-art baseline models on all the metrics under both training paradigms, which demonstrates the effectiveness of our methods that includes the knowledge of references. Compared to KG-KE-KR-M, CopyRNN-\textsc{Gater} achieves the same or better results on all the datasets. This suggests that both the retrieved document and keyphrases are useful for predicting absent keyphrases. 

For present keyphrase prediction, we find that \textsc{Gater} outperforms most of the baseline methods on both training paradigms, 
% \footnote{Note that the present $F_1@5$ metric for \textsc{One2One} and \textsc{One2Seq} paradigm cannot be directly compared, since we appends random wrong answers to the predictions until it reaches 5 predictions, so there are certain advantages for \textsc{One2One} which always has sufficient predictions.}
which indicates that the related references also help the model to understand the source document and to predict more accurate present keyphrases.

\subsection{Ablation Study}
To examine the contribution of each component in \textsc{Gater}, we conduct ablation experiments on the largest dataset \textbf{KP20k}, the results of which are presented in Table \ref{tab:ablation}. For the input references, the model's performance is degraded if either the retrieved documents or retrieved keyphrases are removed, which indicates that both are useful for keyphrases prediction.
% Removing both the retrieved documents and keyphrases means that no references are available. 
For the heterogeneous graph encoder, the graph becomes a heterogeneous bipartite graph when the $d2d$ edges are removed, and a homogeneous graph when the $w2d$ edges are removed. We can see that both result in degraded performance due to the lack of interaction. Removing both the $d2d$ edges and the $w2d$ edges means that the reference information is only used on the decoder side with the reference-aware decoder, which further degrades the results. 
For the reference-aware decoder, we find the hierarchical attention and copy mechanism to be essential to the performance of \textsc{Gater}. 
% If both are removed, the results are slightly less satisfactory than catSeq.
This indicates the importance of integrating knowledge from references on the decoder side.

\begin{table}[t]
\centering
\scalebox{0.7}{
\begin{tabular}{l|cc|cc}
\toprule
\multicolumn{1}{l|}{\multirow{2}{*}{\textbf{Model}}} & \multicolumn{2}{c|}{\textbf{Present}} & \multicolumn{2}{c}{\textbf{Absent}} \\
\multicolumn{1}{c|}{} & $F_1@5$ & $F_1@M$ & $F_1@5$ & $F_1@M$ \\
\hline
catSeq-\textsc{Gater} & \textbf{0.295} & \textbf{0.384} & \textbf{0.030} & \textbf{0.060} \\
\hline
\textit{Input Reference} &  &  & & \\
\quad - retrieved documents &  0.293 & 0.377 & 0.026 & 0.052 \\
\quad - retrieved keyphrases &  0.291 & 0.369 & 0.018 & 0.037 \\
\quad - both & 0.291 & 0.367 & 0.015 & 0.032 \\
\hline
\textit{Heterogeneous Graph Encoder} &  &  &  &  \\
\quad - $d2d$ edge &  0.294 & 0.379 & 0.024 & 0.049 \\
\quad - $w2d$ edge &  0.294 & 0.379 & 0.026 & 0.052 \\
\quad - both &  0.293 & 0.371 & 0.020 & 0.041 \\
\hline
\textit{Reference-aware Decoder} &  &  &  &  \\
\quad - hierarchical copy & 0.293 & 0.373 & 0.022 & 0.042 \\
\quad \quad - hierarchical attention & 0.291 & 0.368 & 0.018 & 0.036 \\
\bottomrule
\end{tabular}}
\caption{Ablation study of catSeq-\textsc{Gater} on \textbf{KP20k} dataset. All references are ignored in graph encoder when removing $d2d$ edge and the heterogeneous graph becomes homogeneous graph when removing $w2d$ edge.}
\label{tab:ablation}
\end{table}

\subsection{Quality and Influence of References}
\label{section:references}

\begin{figure}[t]
\centering
\includegraphics[width=3in]{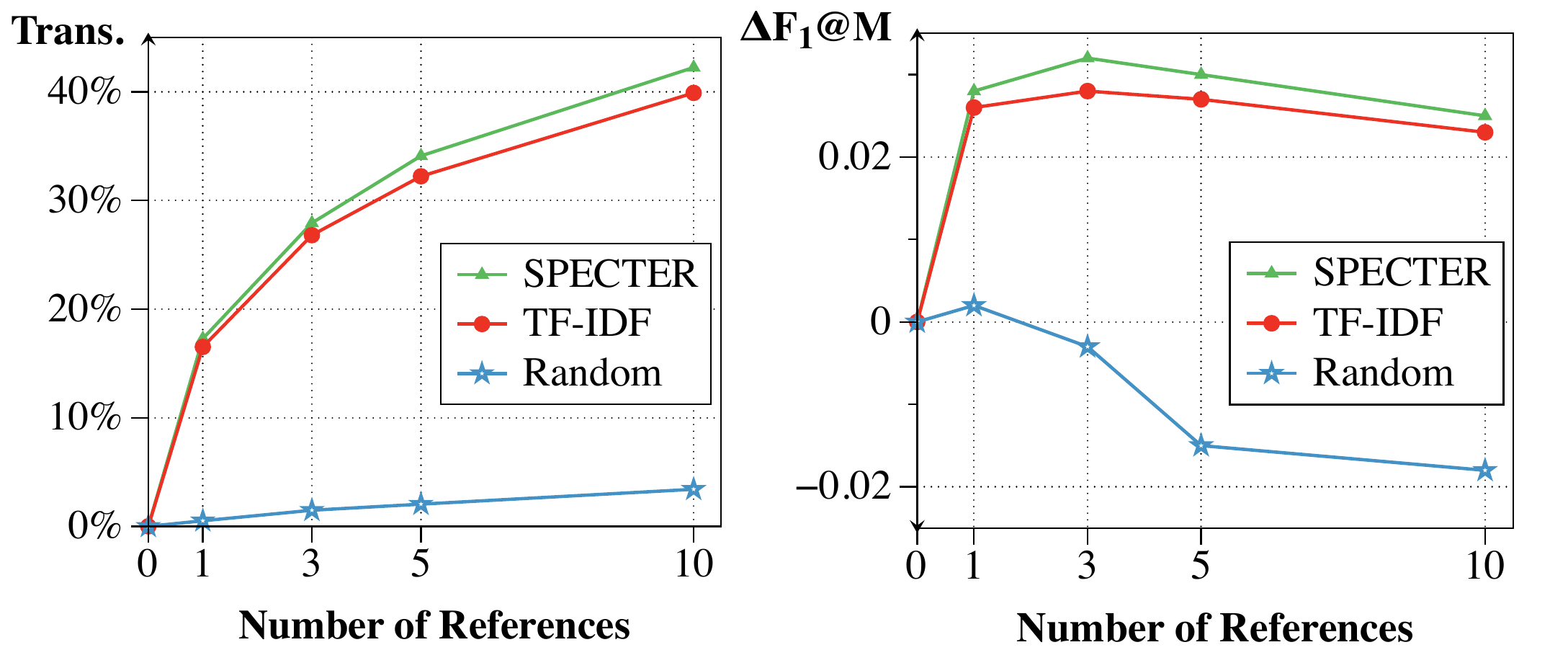}
\caption{Transforming rate and $\Delta F_1@M$ for absent keyphrases under different types of retrievers on \textbf{KP20k} dataset for catSeq-\textsc{Gater}. We study a random retriever, a sparse retriever based on TF-IDF and a dense retriever based on \textsc{Specter}.}
\label{fig:number_of_references}
\end{figure}

As our graph is based on the retrieved references, we also investigated the quality and influence of the references. We define the quality of the retrieved references as the \textit{transforming rate} of absent keyphrase (i.e., the proportion of absent keyphrases that appear in the retrieved references). Intuitively, the references that contain more absent keyphrases provide more explicit knowledge for the model generation. As shown on the left part in Figure \ref{fig:number_of_references}, the simple sparse retriever based on TF-IDF outperforms the random retriever by a large margin regarding the reference quality. We also use a dense retriever \textsc{Specter}\footnote{https://github.com/allenai/specter} \cite{cohan2020specter}, which is a BERT-based model pretrained using scientific documents. We find that using a dense retriever further helps in the transforming rate of absent keyphrases. On the right part of Figure \ref{fig:number_of_references}, we show the influence of the references, and we note that random references degrade the model performance as they contain a lot of noise. Surprisingly, we can obtain a 2.6\% performance boost in the prediction of absent keyphrase by considering only the most similar references with a sparse or dense retriever, and the introduction of more than three references does not further improve the performance. One possible explanation is that although more references lead to a higher transforming rate of the absent keyphrase, they also introduce more irrelevant information, which interferes with the judgment of the model.

\subsection{Incorporating Baselines with \textsc{Gater}}
\label{sec:incorporating}

\begin{table}[t]
\centering
\scalebox{0.7}{
\begin{tabular}{l|cc|cc}
\toprule
\multicolumn{1}{l|}{\multirow{2}{*}{\textbf{Model}}} & \multicolumn{2}{c|}{\textbf{Present}} & \multicolumn{2}{c}{\textbf{Absent}} \\
\multicolumn{1}{c|}{} & $F1@5$ & $F1@M$ & $F1@5$ & $F1@M$ \\
\hline
catSeqD & 0.285 & 0.363 & 0.015 & 0.031 \\
\quad + \textsc{Gater} & \textbf{0.294} & \textbf{0.381} & \textbf{0.025} & \textbf{0.051} \\
\hline
catSeqCorr & 0.289 & 0.365 & 0.015 & 0.032 \\
\quad + \textsc{Gater} & \textbf{0.296} & \textbf{0.384} & \textbf{0.030} & \textbf{0.060} \\
\hline
catSeqTG & 0.292 & 0.366 & 0.015 & 0.032 \\
\quad + \textsc{Gater} & \textbf{0.293} & \textbf{0.380} & \textbf{0.025} & \textbf{0.052} \\
\bottomrule
\end{tabular}}
\caption{Results of applying our \textsc{Gater} to other baseline models on \textbf{KP20k} test set. The best results are bold.}
\label{tab:incorporate_baseline}
\end{table}

Our proposed \textsc{Gater} can be considered as an extra plugin for incorporating knowledge from references on both the encoder and decoder sides, which can also be easily applied to other models. We investigate the effects of adding \textsc{Gater} to other baseline models in Table \ref{tab:incorporate_baseline}. We note that \textsc{Gater} enhances the performance of all the baseline models in both predicting present and absent keyphrases. This further demonstrates the effectiveness and portability of the proposed method.

\begin{figure*}[t]
\centering
\includegraphics[width=6.2in]{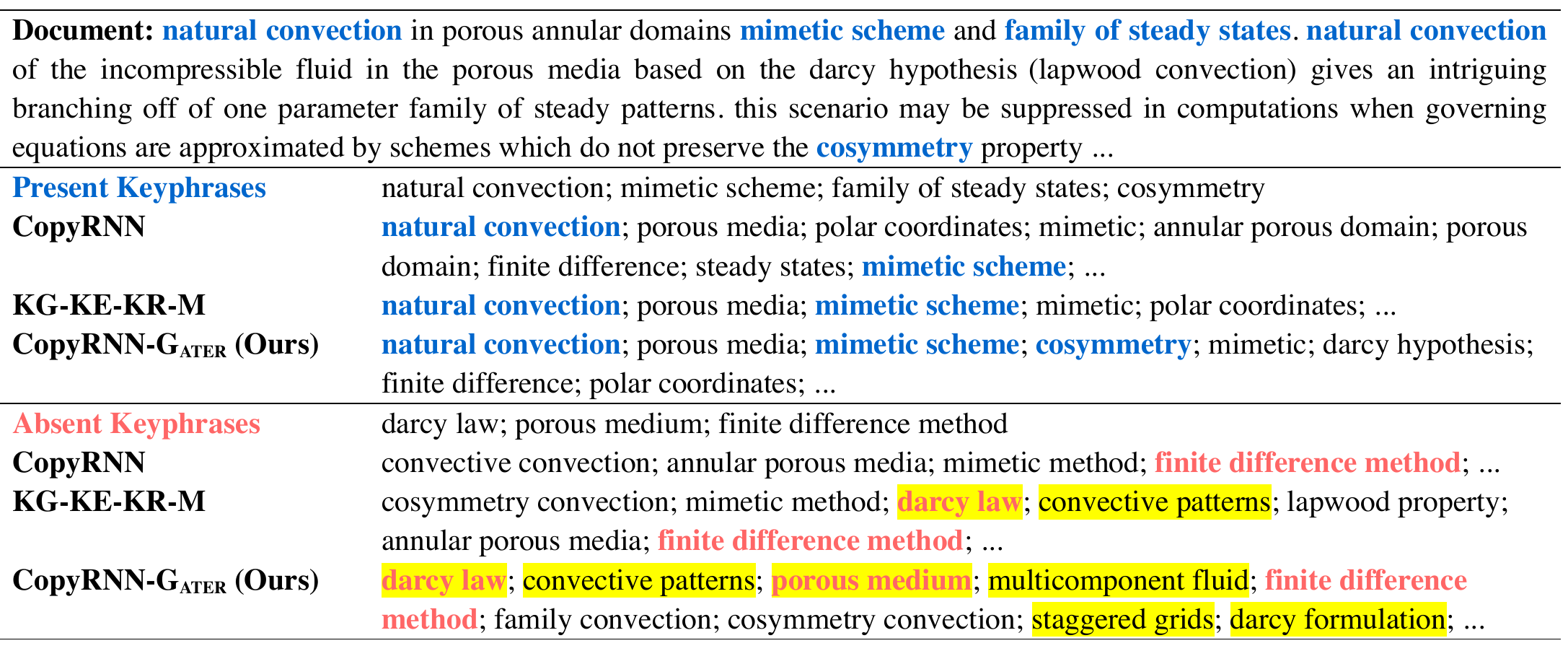}
\caption{Example of generated keyphrases by different models. The top 10 predictions are compared and some incorrect predictions are omitted for simplicity. The correct predictions are in bold blue and bold red for present and absent keyphrase, respectively. The absent predictions that appear in the references are highlighted in yellow, where only the keyphrases of retrieved documents are considered as references for KG-KE-KR-M.}
\label{tab:case_study}
\end{figure*}

\subsection{Case Study}
We display a prediction example by baseline models and CopyRNN-\textsc{Gater} in Figure \ref{tab:case_study}. Our model generates more accurate present and absent keyphrases comparing to the baselines. For instance, we observe that CopyRNN-\textsc{Gater} successfully predicts the absent keyphrase ``porous medium'' as it appears in the retrieved documents, while both CopyRNN and KG-KE-KR-M fail. This demonstrates that using both the retrieved documents and keyphrases as references provides more knowledge (e.g., candidates of the ground-truth absent keyphrases) compared with using keyphrases alone as in KG-KE-KR-M.

\section{Related Work}
\subsection{Keyphrase Extraction and Generation}
Existing approaches for keyphrase prediction can be broadly divided into extraction and generation methods. 
Early work mostly use a two-step approach for keyphrase extraction. 
First, they extract a large set of candidate phrases by hand-crafted rules \cite{mihalcea2004textrank,medelyan2009human,liu2011automatic}. Then, these candidates are scored and reranked based on unsupervised methods \cite{mihalcea2004textrank,wan2008single} or supervised methods \cite{hulth2003improved,nguyen2007keyphrase}. 
Other extractive approaches utilize neural-based sequence labeling methods \cite{zhang2016keyphrase,gollapalli2017incorporating}. 

Keyphrase generation is an extension of keyphrase extraction which considers the absent keyphrase prediction. \citet{meng2017} proposed a generative model CopyRNN based on the encoder-decoder framework \cite{sutskever2014sequence}. They employed an \textsc{One2One} paradigm that uses a single keyphrase as the target sequence.
Since CopyRNN uses beam search to perform independently prediction, it's lack of dependency on the generated keyphrases, which results in many duplicated keyphrases. 
CorrRNN \cite{chen2018a} proposed a review mechanism to consider the hidden states of the previously generated keyphrase. 
\citet{ye2018} proposed to use a seperator $\langle\mathit{sep}\rangle$ to concatnate all keyphrases as a sequence in training. With this setup, the seq2seq model is capable to generate all possible keyphrases in one sequence as well as capture the contextual information between the keyphrases. However, it still use beam search to generate multiple keyphrases sequences with a fixed beam depth, and then perform keyphrase ranking to select top-k keyphrases as output. 
\citet{yuan2018} proposed catSeq with \textsc{One2Seq} paradigm by adding a special token $\langle\mathit{eos}\rangle$ at the end to terminate the decoding process. They further introduce catSeqD by maximizing mutual information between all the keyphrases and source text and using orthogonal constraints \cite{bousmalis2016domain} to ensure the coverage and diversity of the generated keyphrase. 
Many works are conducted based on the \textsc{One2Seq} paradigm \cite{chen2019,chan2019,chen2020a,meng2020empirical,luo2020sensenet}. 
\citet{chen2019} proposed to use the keyphrases of retrieved documents as an external input. However, the keyphrase alone lacks semantic information, and the potential knowledge in the retrieved documents are also ignored. In contrast, our method makes full use of both retrieved documents and keyphrases as references.
Since catSeq tends to generate shorter sequences, a reinforcement learning approach is introduced by \citet{chan2019} to encourage their model to generate the correct number of keyphrases with an adaptive reward (i.e., $F_1$ and $Recall$). 
More recently, \citet{luo2021keyphrase} introduced a two-stage reinforcement learning-based fine-tuning approach with a fine-grained reward score, which also considers the semantic similarities between predictions and targets. \citet{ye2021one2set} proposed a \textsc{One2Set} paradigm to predict the keyphrases as a set, which eliminates the bias caused by the predefined order in \textsc{One2Seq} paradigm. Our method can also be integrated into these methods to further improve performance, as shown in section \ref{sec:incorporating}.

\subsection{Heterogeneous Graph for NLP}
Different from homogeneous graph that only considers a single type of nodes or links, heterogeneous graph can deal with multiple types of nodes or links \cite{shi2016survey}. \citet{linmei2019heterogeneous} constructed a topic-entity heterogeneous neural graph for semi-supervised short text classification. \citet{tu2019multi} introduced a heterogeneous graph neural network to encode documents, entities, and candidates together for multi-hop reading comprehension. \citet{wang2020heterogeneous} presented heterogeneous graph neural network with words, sentences, and documents nodes for extractive summarization. In our paper, we study the keyword-document heterogeneous graph network for keyphrase generation, which has not been explored before.

\section{Conclusions}
In this paper,  we propose a graph-based method that can capture explicit knowledge from related references. Our model consists of a heterogeneous graph encoder to model different granularity of relations among the source document and its references, and a hierarchical attention and copy mechanism to guide the decoding process. Extensive experiments demonstrate the effectiveness and portability of our method on both the present and absent keyphrase predictions.

\section*{Acknowledgments}
The authors wish to thank the anonymous reviewers for their helpful comments. This work was partially funded by China National Key R\&D Program (No. 2017YFB1002104), National Natural Science Foundation of China (No. 61976056, 62076069), Shanghai Municipal Science and Technology Major Project (No.2021SHZDZX0103).

\bibliography{emnlp2021}

\begin{thebibliography}{36}
\expandafter\ifx\csname natexlab\endcsname\relax\def\natexlab#1{#1}\fi

\bibitem[{Bahdanau et~al.(2015)Bahdanau, Cho, and Bengio}]{bahdanau2014neural}
Dzmitry Bahdanau, Kyunghyun Cho, and Yoshua Bengio. 2015.
\newblock \href {http://arxiv.org/abs/1409.0473} {Neural machine translation by
  jointly learning to align and translate}.
\newblock In \emph{3rd International Conference on Learning Representations,
  {ICLR} 2015, San Diego, CA, USA, May 7-9, 2015, Conference Track
  Proceedings}.

\bibitem[{Bousmalis et~al.(2016)Bousmalis, Trigeorgis, Silberman, Krishnan, and
  Erhan}]{bousmalis2016domain}
Konstantinos Bousmalis, George Trigeorgis, Nathan Silberman, Dilip Krishnan,
  and Dumitru Erhan. 2016.
\newblock \href
  {https://proceedings.neurips.cc/paper/2016/hash/45fbc6d3e05ebd93369ce542e8f2322d-Abstract.html}
  {Domain separation networks}.
\newblock In \emph{Advances in Neural Information Processing Systems 29: Annual
  Conference on Neural Information Processing Systems 2016, December 5-10,
  2016, Barcelona, Spain}, pages 343--351.

\bibitem[{Chan et~al.(2019)Chan, Chen, Wang, and King}]{chan2019}
Hou~Pong Chan, Wang Chen, Lu~Wang, and Irwin King. 2019.
\newblock \href {https://doi.org/10.18653/v1/P19-1208} {Neural keyphrase
  generation via reinforcement learning with adaptive rewards}.
\newblock In \emph{Proceedings of the 57th Annual Meeting of the Association
  for Computational Linguistics}, pages 2163--2174, Florence, Italy.
  Association for Computational Linguistics.

\bibitem[{Chen et~al.(2018)Chen, Zhang, Wu, Yan, and Li}]{chen2018a}
Jun Chen, Xiaoming Zhang, Yu~Wu, Zhao Yan, and Zhoujun Li. 2018.
\newblock \href {https://doi.org/10.18653/v1/D18-1439} {Keyphrase generation
  with correlation constraints}.
\newblock In \emph{Proceedings of the 2018 Conference on Empirical Methods in
  Natural Language Processing}, pages 4057--4066, Brussels, Belgium.
  Association for Computational Linguistics.

\bibitem[{Chen et~al.(2019{\natexlab{a}})Chen, Chan, Li, Bing, and
  King}]{chen2019}
Wang Chen, Hou~Pong Chan, Piji Li, Lidong Bing, and Irwin King.
  2019{\natexlab{a}}.
\newblock \href {https://doi.org/10.18653/v1/N19-1292} {An integrated approach
  for keyphrase generation via exploring the power of retrieval and
  extraction}.
\newblock In \emph{Proceedings of the 2019 Conference of the North {A}merican
  Chapter of the Association for Computational Linguistics: Human Language
  Technologies, Volume 1 (Long and Short Papers)}, pages 2846--2856,
  Minneapolis, Minnesota. Association for Computational Linguistics.

\bibitem[{Chen et~al.(2020)Chen, Chan, Li, and King}]{chen2020a}
Wang Chen, Hou~Pong Chan, Piji Li, and Irwin King. 2020.
\newblock \href {https://doi.org/10.18653/v1/2020.acl-main.103} {Exclusive
  hierarchical decoding for deep keyphrase generation}.
\newblock In \emph{Proceedings of the 58th Annual Meeting of the Association
  for Computational Linguistics}, pages 1095--1105, Online. Association for
  Computational Linguistics.

\bibitem[{Chen et~al.(2019{\natexlab{b}})Chen, Gao, Zhang, King, and
  Lyu}]{chen2019a}
Wang Chen, Yifan Gao, Jiani Zhang, Irwin King, and Michael~R. Lyu.
  2019{\natexlab{b}}.
\newblock \href {https://doi.org/10.1609/aaai.v33i01.33016268} {Title-guided
  encoding for keyphrase generation}.
\newblock In \emph{The Thirty-Third {AAAI} Conference on Artificial
  Intelligence, {AAAI} 2019, The Thirty-First Innovative Applications of
  Artificial Intelligence Conference, {IAAI} 2019, The Ninth {AAAI} Symposium
  on Educational Advances in Artificial Intelligence, {EAAI} 2019, Honolulu,
  Hawaii, USA, January 27 - February 1, 2019}, pages 6268--6275. {AAAI} Press.

\bibitem[{Cho et~al.(2014)Cho, van Merri{\"e}nboer, Gulcehre, Bahdanau,
  Bougares, Schwenk, and Bengio}]{cho2014learning}
Kyunghyun Cho, Bart van Merri{\"e}nboer, Caglar Gulcehre, Dzmitry Bahdanau,
  Fethi Bougares, Holger Schwenk, and Yoshua Bengio. 2014.
\newblock \href {https://doi.org/10.3115/v1/D14-1179} {Learning phrase
  representations using {RNN} encoder{--}decoder for statistical machine
  translation}.
\newblock In \emph{Proceedings of the 2014 Conference on Empirical Methods in
  Natural Language Processing ({EMNLP})}, pages 1724--1734, Doha, Qatar.
  Association for Computational Linguistics.

\bibitem[{Cohan et~al.(2020)Cohan, Feldman, Beltagy, Downey, and
  Weld}]{cohan2020specter}
Arman Cohan, Sergey Feldman, Iz~Beltagy, Doug Downey, and Daniel Weld. 2020.
\newblock \href {https://doi.org/10.18653/v1/2020.acl-main.207} {{SPECTER}:
  Document-level representation learning using citation-informed transformers}.
\newblock In \emph{Proceedings of the 58th Annual Meeting of the Association
  for Computational Linguistics}, pages 2270--2282, Online. Association for
  Computational Linguistics.

\bibitem[{Gollapalli et~al.(2017)Gollapalli, Li, and
  Yang}]{gollapalli2017incorporating}
Sujatha~Das Gollapalli, Xiaoli Li, and Peng Yang. 2017.
\newblock \href {http://aaai.org/ocs/index.php/AAAI/AAAI17/paper/view/14628}
  {Incorporating expert knowledge into keyphrase extraction}.
\newblock In \emph{Proceedings of the Thirty-First {AAAI} Conference on
  Artificial Intelligence, February 4-9, 2017, San Francisco, California,
  {USA}}, pages 3180--3187. {AAAI} Press.

\bibitem[{Gu et~al.(2016)Gu, Lu, Li, and Li}]{gu2016incorporating}
Jiatao Gu, Zhengdong Lu, Hang Li, and Victor~O.K. Li. 2016.
\newblock \href {https://doi.org/10.18653/v1/P16-1154} {Incorporating copying
  mechanism in sequence-to-sequence learning}.
\newblock In \emph{Proceedings of the 54th Annual Meeting of the Association
  for Computational Linguistics (Volume 1: Long Papers)}, pages 1631--1640,
  Berlin, Germany. Association for Computational Linguistics.

\bibitem[{Hulth(2003)}]{hulth2003improved}
Anette Hulth. 2003.
\newblock Improved automatic keyword extraction given more linguistic
  knowledge.
\newblock In \emph{Proceedings of the 2003 conference on Empirical methods in
  natural language processing}, pages 216--223.

\bibitem[{Kim et~al.(2010)Kim, Medelyan, Kan, and Baldwin}]{kim2010semeval}
Su~Nam Kim, Olena Medelyan, Min-Yen Kan, and Timothy Baldwin. 2010.
\newblock \href {https://www.aclweb.org/anthology/S10-1004} {{S}em{E}val-2010
  task 5 : Automatic keyphrase extraction from scientific articles}.
\newblock In \emph{Proceedings of the 5th International Workshop on Semantic
  Evaluation}, pages 21--26, Uppsala, Sweden. Association for Computational
  Linguistics.

\bibitem[{Krapivin et~al.(2009)Krapivin, Autaeu, and
  Marchese}]{krapivin2009large}
Mikalai Krapivin, Aliaksandr Autaeu, and Maurizio Marchese. 2009.
\newblock Large dataset for keyphrases extraction.
\newblock Technical report, University of Trento.

\bibitem[{Linmei et~al.(2019)Linmei, Yang, Shi, Ji, and
  Li}]{linmei2019heterogeneous}
Hu~Linmei, Tianchi Yang, Chuan Shi, Houye Ji, and Xiaoli Li. 2019.
\newblock \href {https://doi.org/10.18653/v1/D19-1488} {Heterogeneous graph
  attention networks for semi-supervised short text classification}.
\newblock In \emph{Proceedings of the 2019 Conference on Empirical Methods in
  Natural Language Processing and the 9th International Joint Conference on
  Natural Language Processing (EMNLP-IJCNLP)}, pages 4821--4830, Hong Kong,
  China. Association for Computational Linguistics.

\bibitem[{Liu et~al.(2011)Liu, Chen, Zheng, and Sun}]{liu2011automatic}
Zhiyuan Liu, Xinxiong Chen, Yabin Zheng, and Maosong Sun. 2011.
\newblock \href {https://www.aclweb.org/anthology/W11-0316} {Automatic
  keyphrase extraction by bridging vocabulary gap}.
\newblock In \emph{Proceedings of the Fifteenth Conference on Computational
  Natural Language Learning}, pages 135--144, Portland, Oregon, USA.
  Association for Computational Linguistics.

\bibitem[{Luo et~al.(2020)Luo, Li, Wang, Xing, Zhang, and
  Huang}]{luo2020sensenet}
Yichao Luo, Zhengyan Li, Bingning Wang, Xiaoyu Xing, Qi~Zhang, and Xuanjing
  Huang. 2020.
\newblock Sensenet: Neural keyphrase generation with document structure.
\newblock In \emph{arXiv}.

\bibitem[{Luo et~al.(2021)Luo, Xu, Ye, Qiu, and Zhang}]{luo2021keyphrase}
Yichao Luo, Yige Xu, Jiacheng Ye, Xipeng Qiu, and Qi~Zhang. 2021.
\newblock Keyphrase generation with fine-grained evaluation-guided
  reinforcement learning.
\newblock \emph{arXiv preprint arXiv:2104.08799}.

\bibitem[{Luong et~al.(2015)Luong, Pham, and
  Manning}]{luong-etal-2015-effective}
Thang Luong, Hieu Pham, and Christopher~D. Manning. 2015.
\newblock \href {https://doi.org/10.18653/v1/D15-1166} {Effective approaches to
  attention-based neural machine translation}.
\newblock In \emph{Proceedings of the 2015 Conference on Empirical Methods in
  Natural Language Processing}, pages 1412--1421, Lisbon, Portugal. Association
  for Computational Linguistics.

\bibitem[{Medelyan et~al.(2009)Medelyan, Frank, and Witten}]{medelyan2009human}
Olena Medelyan, Eibe Frank, and Ian~H. Witten. 2009.
\newblock \href {https://www.aclweb.org/anthology/D09-1137} {Human-competitive
  tagging using automatic keyphrase extraction}.
\newblock In \emph{Proceedings of the 2009 Conference on Empirical Methods in
  Natural Language Processing}, pages 1318--1327, Singapore. Association for
  Computational Linguistics.

\bibitem[{Meng et~al.(2021)Meng, Yuan, Wang, Zhao, Trischler, and
  He}]{meng2020empirical}
Rui Meng, Xingdi Yuan, Tong Wang, Sanqiang Zhao, Adam Trischler, and Daqing He.
  2021.
\newblock \href {https://www.aclweb.org/anthology/2021.naacl-main.396} {An
  empirical study on neural keyphrase generation}.
\newblock In \emph{Proceedings of the 2021 Conference of the North American
  Chapter of the Association for Computational Linguistics: Human Language
  Technologies}, pages 4985--5007, Online. Association for Computational
  Linguistics.

\bibitem[{Meng et~al.(2017)Meng, Zhao, Han, He, Brusilovsky, and
  Chi}]{meng2017}
Rui Meng, Sanqiang Zhao, Shuguang Han, Daqing He, Peter Brusilovsky, and
  Yu~Chi. 2017.
\newblock \href {https://doi.org/10.18653/v1/P17-1054} {Deep keyphrase
  generation}.
\newblock In \emph{Proceedings of the 55th Annual Meeting of the Association
  for Computational Linguistics (Volume 1: Long Papers)}, pages 582--592,
  Vancouver, Canada. Association for Computational Linguistics.

\bibitem[{Mihalcea and Tarau(2004)}]{mihalcea2004textrank}
Rada Mihalcea and Paul Tarau. 2004.
\newblock \href {https://www.aclweb.org/anthology/W04-3252} {{T}ext{R}ank:
  Bringing order into text}.
\newblock In \emph{Proceedings of the 2004 Conference on Empirical Methods in
  Natural Language Processing}, pages 404--411, Barcelona, Spain. Association
  for Computational Linguistics.

\bibitem[{Nguyen and Kan(2007)}]{nguyen2007keyphrase}
Thuy~Dung Nguyen and Min-Yen Kan. 2007.
\newblock Keyphrase extraction in scientific publications.
\newblock In \emph{International conference on Asian digital libraries}.

\bibitem[{See et~al.(2017)See, Liu, and Manning}]{see-etal-2017-get}
Abigail See, Peter~J. Liu, and Christopher~D. Manning. 2017.
\newblock \href {https://doi.org/10.18653/v1/P17-1099} {Get to the point:
  Summarization with pointer-generator networks}.
\newblock In \emph{Proceedings of the 55th Annual Meeting of the Association
  for Computational Linguistics (Volume 1: Long Papers)}, pages 1073--1083,
  Vancouver, Canada. Association for Computational Linguistics.

\bibitem[{Shi et~al.(2016)Shi, Li, Zhang, Sun, and Philip}]{shi2016survey}
Chuan Shi, Yitong Li, Jiawei Zhang, Yizhou Sun, and S~Yu Philip. 2016.
\newblock A survey of heterogeneous information network analysis.
\newblock In \emph{TKDE}.

\bibitem[{Sutskever et~al.(2014)Sutskever, Vinyals, and
  Le}]{sutskever2014sequence}
Ilya Sutskever, Oriol Vinyals, and Quoc~V. Le. 2014.
\newblock \href
  {https://proceedings.neurips.cc/paper/2014/hash/a14ac55a4f27472c5d894ec1c3c743d2-Abstract.html}
  {Sequence to sequence learning with neural networks}.
\newblock In \emph{Advances in Neural Information Processing Systems 27: Annual
  Conference on Neural Information Processing Systems 2014, December 8-13 2014,
  Montreal, Quebec, Canada}, pages 3104--3112.

\bibitem[{Tu et~al.(2019)Tu, Wang, Huang, Tang, He, and Zhou}]{tu2019multi}
Ming Tu, Guangtao Wang, Jing Huang, Yun Tang, Xiaodong He, and Bowen Zhou.
  2019.
\newblock \href {https://doi.org/10.18653/v1/P19-1260} {Multi-hop reading
  comprehension across multiple documents by reasoning over heterogeneous
  graphs}.
\newblock In \emph{Proceedings of the 57th Annual Meeting of the Association
  for Computational Linguistics}, pages 2704--2713, Florence, Italy.
  Association for Computational Linguistics.

\bibitem[{Vaswani et~al.(2017)Vaswani, Shazeer, Parmar, Uszkoreit, Jones,
  Gomez, Kaiser, and Polosukhin}]{vaswani2017attention}
Ashish Vaswani, Noam Shazeer, Niki Parmar, Jakob Uszkoreit, Llion Jones,
  Aidan~N. Gomez, Lukasz Kaiser, and Illia Polosukhin. 2017.
\newblock \href
  {https://proceedings.neurips.cc/paper/2017/hash/3f5ee243547dee91fbd053c1c4a845aa-Abstract.html}
  {Attention is all you need}.
\newblock In \emph{Advances in Neural Information Processing Systems 30: Annual
  Conference on Neural Information Processing Systems 2017, December 4-9, 2017,
  Long Beach, CA, {USA}}, pages 5998--6008.

\bibitem[{Velickovic et~al.(2018)Velickovic, Cucurull, Casanova, Romero,
  Li{\`{o}}, and Bengio}]{velivckovic2018graph}
Petar Velickovic, Guillem Cucurull, Arantxa Casanova, Adriana Romero, Pietro
  Li{\`{o}}, and Yoshua Bengio. 2018.
\newblock \href {https://openreview.net/forum?id=rJXMpikCZ} {Graph attention
  networks}.
\newblock In \emph{6th International Conference on Learning Representations,
  {ICLR} 2018, Vancouver, BC, Canada, April 30 - May 3, 2018, Conference Track
  Proceedings}. OpenReview.net.

\bibitem[{Wan and Xiao(2008)}]{wan2008single}
Xiaojun Wan and Jianguo Xiao. 2008.
\newblock Single document keyphrase extraction using neighborhood knowledge.
\newblock In \emph{AAAI}.

\bibitem[{Wang et~al.(2020)Wang, Liu, Zheng, Qiu, and
  Huang}]{wang2020heterogeneous}
Danqing Wang, Pengfei Liu, Yining Zheng, Xipeng Qiu, and Xuanjing Huang. 2020.
\newblock \href {https://doi.org/10.18653/v1/2020.acl-main.553} {Heterogeneous
  graph neural networks for extractive document summarization}.
\newblock In \emph{Proceedings of the 58th Annual Meeting of the Association
  for Computational Linguistics}, pages 6209--6219, Online. Association for
  Computational Linguistics.

\bibitem[{Ye and Wang(2018)}]{ye2018}
Hai Ye and Lu~Wang. 2018.
\newblock \href {https://doi.org/10.18653/v1/D18-1447} {Semi-supervised
  learning for neural keyphrase generation}.
\newblock In \emph{Proceedings of the 2018 Conference on Empirical Methods in
  Natural Language Processing}, pages 4142--4153, Brussels, Belgium.
  Association for Computational Linguistics.

\bibitem[{Ye et~al.(2021)Ye, Gui, Luo, Xu, and Zhang}]{ye2021one2set}
Jiacheng Ye, Tao Gui, Yichao Luo, Yige Xu, and Qi~Zhang. 2021.
\newblock \href {https://doi.org/10.18653/v1/2021.acl-long.354} {{O}ne2{S}et:
  {G}enerating diverse keyphrases as a set}.
\newblock In \emph{Proceedings of the 59th Annual Meeting of the Association
  for Computational Linguistics and the 11th International Joint Conference on
  Natural Language Processing (Volume 1: Long Papers)}, pages 4598--4608,
  Online. Association for Computational Linguistics.

\bibitem[{Yuan et~al.(2020)Yuan, Wang, Meng, Thaker, Brusilovsky, He, and
  Trischler}]{yuan2018}
Xingdi Yuan, Tong Wang, Rui Meng, Khushboo Thaker, Peter Brusilovsky, Daqing
  He, and Adam Trischler. 2020.
\newblock \href {https://doi.org/10.18653/v1/2020.acl-main.710} {One size does
  not fit all: Generating and evaluating variable number of keyphrases}.
\newblock In \emph{Proceedings of the 58th Annual Meeting of the Association
  for Computational Linguistics}, pages 7961--7975, Online. Association for
  Computational Linguistics.

\bibitem[{Zhang et~al.(2016)Zhang, Wang, Gong, and Huang}]{zhang2016keyphrase}
Qi~Zhang, Yang Wang, Yeyun Gong, and Xuanjing Huang. 2016.
\newblock \href {https://doi.org/10.18653/v1/D16-1080} {Keyphrase extraction
  using deep recurrent neural networks on {T}witter}.
\newblock In \emph{Proceedings of the 2016 Conference on Empirical Methods in
  Natural Language Processing}, pages 836--845, Austin, Texas. Association for
  Computational Linguistics.

\end{thebibliography}
\bibliographystyle{acl_natbib}
\newpage

\appendix

\section{Results on Krapivin Dataset}
\label{sec:appendix}

\begin{table}[h]
\centering
\scalebox{0.65}{
\begin{tabular}{l|llll}
\toprule
\multicolumn{1}{l|}{\multirow{3}{*}{\textbf{Model}}} & \multicolumn{4}{c}{\textbf{Krapivin}}  \\

& \multicolumn{2}{c}{\textbf{Present}} & \multicolumn{2}{c}{\textbf{Absent}} \\

\multicolumn{1}{c|}{} & {$F1@5$} & $F1@10$ & $R@10$ & $R@50$ \\
\hline
CopyRNN \citep{meng2017} & 0.334 & 0.326 & 0.113 & 0.202 \\
CorrRNN \citep{chen2018a} & 0.358 & 0.330 & 0.108 & - \\
TG-Net \citep{chen2019a}  & 0.406 & 0.370 & 0.146 & 0.253 \\
KG-KE-KR-M \citep{chen2019} & 0.431 & 0.378 & 0.153 & 0.251 \\
\hline
CopyRNN-\textsc{Gater} (Ours) & \textbf{0.435$_3$} & \textbf{0.383$_2$} & \textbf{0.195$_3$} & \textbf{0.294$_3$} \\
% \% gain & -1\% & 6\% & 72\% & 68\% \\
\bottomrule 

\multicolumn{5}{}{} \\

\toprule
\multicolumn{1}{l|}{\multirow{3}{*}{\textbf{Model}}} & \multicolumn{4}{c}{\textbf{Krapivin}}  \\

& \multicolumn{2}{c}{\textbf{Present}} & \multicolumn{2}{c}{\textbf{Absent}} \\

\multicolumn{1}{c|}{} & $F1@5$ & $F1@M$ & $F1@5$ & $F1@M$ \\
 \hline
catSeq \citep{yuan2018}  & 0.269 & 0.354 & 0.018 & 0.036 \\
catSeqD \citep{yuan2018} & 0.264 & 0.349 & 0.018 & 0.037 \\
catSeqCorr \citep{chan2019} & 0.265 & 0.349 & 0.020 & 0.038 \\
catSeqTG \citep{chan2019} & \textbf{0.282} & 0.366 & 0.018 & 0.034 \\
SenSeNet \citep{luo2020sensenet} & 0.279 & 0.354 & 0.024 & 0.046 \\
\hline
catSeq-\textsc{Gater} (Ours) & 0.276$_3$ & \textbf{0.376$_4$} & \textbf{0.037$_3$} & \textbf{0.069$_5$} \\
% \% gain & -1\% & 6\% & 72\% & 68\% \\
\bottomrule
\end{tabular}
}
\caption{Keyphrase prediction results of the models trained under \textsc{One2One} and \textsc{One2Seq} paradigms. The best results are bold. The subscripts are the corresponding standard deviation (e.g., 0.069$_5$ means 0.069$\pm$0.005).}
\end{table}

\end{document}